\documentclass[11pt]{article}
\usepackage[preprint]{acl} 
\usepackage{times}
\usepackage{latexsym}
\usepackage[T1]{fontenc}
\usepackage[utf8]{inputenc}
\usepackage{microtype}
\usepackage{inconsolata}
\usepackage{graphicx}
\usepackage{amsmath}
\usepackage{booktabs}
\usepackage{multirow}
\usepackage{xcolor}
\usepackage{colortbl}
\usepackage{array}

\usepackage{todonotes}
\usepackage{mathtools}
\usepackage{amsfonts}

\usepackage{graphicx} 
\usepackage{amsmath}
\usepackage{booktabs}
\usepackage{array}
\usepackage[most]{tcolorbox}
\usepackage{tabularx}
\usepackage{longtable}
\usepackage{caption}
\usepackage{wrapfig}
\usepackage{bbm}
\usepackage{multirow}
\usepackage{multicol}
\usepackage{enumitem}
\usepackage{hyperref}

\usepackage{amsthm}
\newtheorem{theorem}{Theorem}

\newtheorem{assumption}[theorem]{Assumption}
\theoremstyle{definition}
\newtheorem{definition}[theorem]{Definition}

\title{The Masked Advantage: Uncovering Local-Language Access to Cultural Knowledge in LLMs}

\author{
 \textbf{Yang Zhang\textsuperscript{1}$^\dagger$},
 \textbf{Xiao Fei\textsuperscript{1}},
 \textbf{Amr Mohamed\textsuperscript{1,2}},
 \\
 \textbf{Sarah Almeida Carneiro\textsuperscript{1}},
 \textbf{Mersin Konomi\textsuperscript{2}},
 \textbf{Mingmeng Geng\textsuperscript{3}},
 \textbf{Ahmed Asaad\textsuperscript{4}},
\\
 \textbf{Guokan Shang\textsuperscript{2}},
 \textbf{Michalis Vazirgiannis\textsuperscript{1,2}$^\dagger$}
\\
\\
 \textsuperscript{1}Ecole Polytechnique,
 \textsuperscript{2}MBZUAI,
 \textsuperscript{3}ENS-PSL,
 \textsuperscript{4}Durham University
\\
 \small{
   $^\dagger$Correspondence: \texttt{yang.zhang@polytechnique.edu, mvazirg@lix.polytechnique.fr}
 }
}
 
\begin{document}
\maketitle

\begin{abstract}
Large language models are increasingly used to answer culturally grounded questions across languages, yet it remains unclear whether local cultural knowledge is better accessed through English or the local language. Existing evaluations face two key limitations: many rely on parallel template-based questions that may not reflect how cultural knowledge naturally appears, and raw accuracy conflates general language proficiency with language-conditioned knowledge access. We address these issues with a controlled framework built on real-world cultural questions collected from regional benchmarks and local sources. By crossing question type (culture-agnostic vs. culture-specific) with query language (English vs. local language), and estimating ability with a shared 1PL item response theory model, we separate proficiency from localized knowledge access. Across 13 locales and roughly 80 models, we find a consistent English advantage on culture-agnostic questions, indicating stronger English proficiency. However, after accounting for this proficiency gap, local languages show a positive knowledge-access advantage in nearly all locale–model settings. This advantage is often masked in raw accuracy but becomes more visible for frontier, regionally aligned, or language-adapted models. Our results suggest that weaker local-language performance does not necessarily imply weaker cultural knowledge; rather, local cultural knowledge may be more accessible through the local language but hidden by limited language proficiency.

\end{abstract}

\section{Introduction}

\begin{figure}[h]
  \includegraphics[width=\columnwidth]{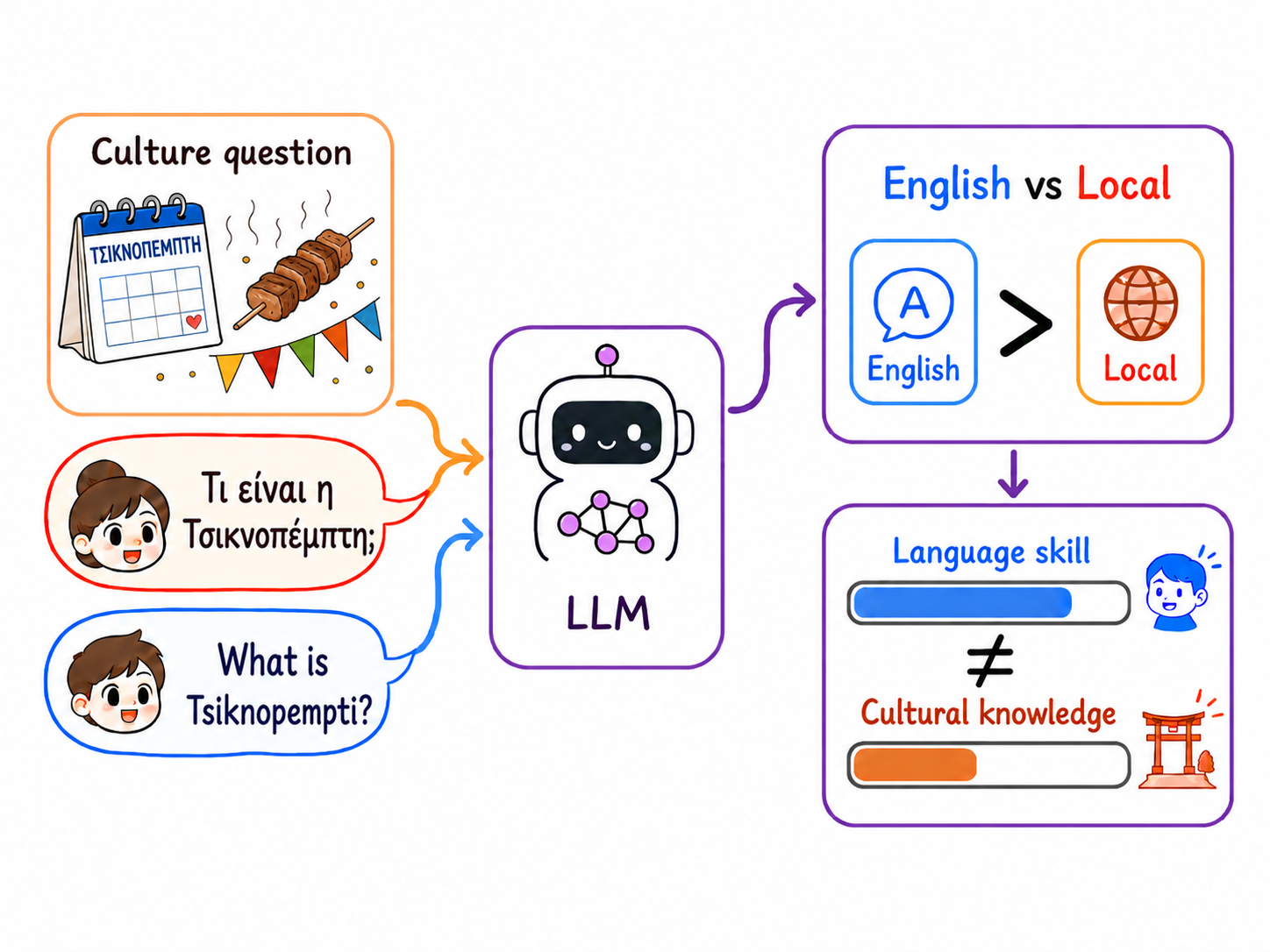}
  \caption{For cultural questions, local knowledge may be masked by weaker local-language proficiency.}
  \label{fig:main_figure}
\end{figure}

Large language models are increasingly regarded as multilingual systems, capable of answering questions written in different languages and concerning history, law, customs, and everyday life across diverse cultural contexts. 
However, the relevant cultural knowledge is often unevenly distributed across languages: some content is produced and discussed primarily in a local language, while training data and evaluation benchmarks remain dominated by English --- the global language. 
This raises two linked questions: \textit{Q1: which query language provides better access to local cultural knowledge?} and \textit{Q2: whether that knowledge is preferentially accessed through the language in which it is grounded?}

Prior work has yielded mixed findings. Several studies suggest that LLMs remain strongly English-centric: knowledge transfers asymmetrically across languages, and translating into English can be a competitive querying strategy, especially for lower-resource languages \citep{yin-etal-2022-geomlama, liu-etal-2025-translation, romanou2025include}.
Another line reports that local-language prompting can help when the query language aligns with the cultural context, and that different languages may activate different knowledge in the same model \citep{ying-etal-2025-disentangling, agarwal2025language, jain2026language,
myung2024blend}.

We argue this disagreement stems partly from how cultural knowledge is measured. Many cultural benchmarks build parallel questions from human-defined templates across languages, which may not reflect how local knowledge naturally appears in each language \citep{ying-etal-2025-disentangling}.
Even with faithful benchmarks, most cross-lingual comparisons rely on raw accuracy, which conflates language proficiency with access to language-conditioned knowledge—two effects that may work in opposite directions. A model may score higher in English simply because it understands English better, while a local language may better activate culturally grounded knowledge yet still yield lower accuracy due to weaker proficiency. A raw English-versus-local gap therefore reports only the net sign of these factors, leaving open which language truly provides better access to local knowledge.


To separate these effects, we compare English and local-language performance on both culture-agnostic and culture-specific questions. The culture-agnostic gap estimates general language proficiency (\textsc{GlobalGap}), while the culture-specific gap captures both proficiency and cultural knowledge access (\textsc{LocalGap}). Their difference gives \textsc{KnowledgeGap}, which isolates language-conditioned knowledge access. 

However, because culture-agnostic and culture-specific items are drawn from different question pools of different intrinsic difficulties, raw accuracy gaps across the two sets are not directly comparable. A lower accuracy on culture-specific questions may simply reflect harder items rather than weaker knowledge access. Therefore, we estimate all gaps with a shared 1PL IRT model \citep{irt, irt4ml}, which jointly estimates item difficulty and model ability on a common scale, enabling the comparison by subtraction. 


We apply this framework across 13 locales and roughly 80 models, using culture-specific items from established regional benchmarks rather than synthetic templates. Our results support the masked advantage hypothesis: models show a consistent English proficiency advantage on culture-agnostic questions, while local-language performance on culture-specific questions is mixed. After controlling for proficiency, however, \textsc{KnowledgeGap} is positive in nearly every locale--model cell, including many where \textsc{LocalGap} is negative. This suggests that local languages often provide better access to cultural knowledge, but this advantage is hidden by weaker local-language proficiency.

Our contributions are threefold:

\begin{enumerate}
    \item We introduce a measurement framework that separates language
    proficiency from language-conditioned cultural knowledge access, casting
    all three gaps as ability contrasts under a single 1PL (Rasch) IRT model
    that accounts for difficulty differences between culture-agnostic and
    culture-specific items.
    \item We conduct a large-scale empirical study across 13 local languages and
    multiple model categories using established benchmarks rather than
    synthetic probes.
    \item We provide evidence for a masked local-language knowledge advantage that is present in nearly all settings but surfaces in raw accuracy only when proficiency does not work against it. Weaker local-language performance therefore appears to stem largely from limited proficiency in that language rather than from missing cultural knowledge itself, with implications for how multilingual LLMs are evaluated and deployed.

\end{enumerate}

\section{Related Work}
\label{sec:related}

\paragraph{Multilingual, Cultural, and Regional Evaluation Benchmarks.} A growing body of benchmarks evaluates LLMs beyond English. Translation-based suites such as Global-MMLU \citep{globalmmlu} show that multilingual evaluation can still inherit English-centric cultural assumptions. While native or regional benchmarks such as INCLUDE \citep{romanou2025include}, CMMLU \citep{li-etal-2024-cmmlu}, ArabicMMLU \citep{koto-etal-2024-arabicmmlu}, KMMLU \citep{son-etal-2025-kmmlu}, MILU \citep{verma-etal-2025-milu}, and GreekMMLU \citep{zhang2026greekmmlu} collect questions from local exams, institutions, and knowledge sources. In parallel, cultural benchmarks such as BLEnD \citep{myung2024blend} extend evaluation from academic knowledge to everyday culture but with a predefined template. There are also works evaluating LLMs under code-switched test, revealing mixed effect of English and local language \citep{mohamed2025lost}.
These benchmarks establish that a language gap exists, but report raw accuracy without separating language proficiency from knowledge access; our framework supplies that decomposition.


\paragraph{Disentangling Language from Knowledge.} 
Closest to our setting, \citet{ying-etal-2025-disentangling} separates language from cultural context and finds that local-language prompts generally improve cultural knowledge access, but it relies on parallel template-based cultural questions, which may not capture how localized knowledge naturally appears in native data. Building on this direction, \citet{agarwal2025language} further asks whether some knowledge is more accessible in a specific language than in English, while \citet{jain2026language} argues that changing the query language can also shift the cultural assumptions activated by the model.
We instead collected real-world cultural questions and cast the problem as a difference-in-differences of model ability estimated by a Bayesian item-response model \citep{irt, equallyinformative}. This lets us recover a knowledge advantage that raw accuracy masks.


\paragraph{IRT for LLM MCQ Benchmarking}

The community has witnessed an increasing number of works applying Item Response Theory (IRT) to LLM Multi-Choice-Question (MCQ) benchmarking, enabling analysis of LLM ability separately from item difficulty. Both tinyBenchmarks \citep{tinybenchmark} and metabench \citep{metabench} use IRT on LLM responses across MCQs to distill benchmarks into much smaller sets, while variants such as PSN-IRT \citep{lostinbenchmarks} and $\beta^3$-IRT \citep{beta3irt} have previously been proposed to extend the capacity of vanilla IRT for more detailed analysis of item difficulty and discrimination. Some recent studies further examine the fitting quality of IRT models with Computerized Adaptive Testing (CAT) \citep{catirt1, catirt2}. However, very few of them have carefully examined the translation and scale invariance problem of logistic IRT models, which is crucial in our particular case of estimating independent model ability on different subsets of the benchmark.

\section{Methodology}
\label{sec:method}

\subsection{Problem Formulation}
\label{sec:formulation}

Let $j$ be a language model. We study two subsets of problems $s\in\{\text{CA},\text{CS}\}$, where culture-specific questions $Q_{\text{CS}}$ target a region with local language $k_{local}$,
alongside culture-agnostic questions $Q_{\text{CA}}$ that require no local
knowledge. Each subset is posed both in $k_{local}$ and in English $k=\text{en}$, so that
the effect of query language is measured within a fixed question pool.

We define all gaps in terms of model ability $\theta_{jsk}$ rather than
raw accuracy, because comparing accuracy across question sets confounds the
effect of interest with differences in their difficulty---a concern for our
central contrast, which compares the disjoint pools $Q_{\text{CS}}$ and
$Q_{\text{CA}}$. We estimate $\theta_{jsk}$, the ability of LLM $j$ on
question set $s$ posed in language $k$, with an Item-Response Model that
places item difficulty and model ability on a common scale
(\S\ref{sec:metrics:crm}).

Our first goal is to examine the \emph{LocalGap}, the ability difference
between querying in $k_{local}$ and in English on culture-specific questions,
\begin{equation}
\text{LocalGap}_j = \theta_{j,\text{CS},k_{local}} - \theta_{j,\text{CS},\text{en}}.
\end{equation}

Its sign alone, however, does not reveal \emph{why} one language wins,
because it conflates \emph{language proficiency}---general ability in
$k_{local}$ versus English---with \emph{language-conditioned knowledge
access}---whether culturally specific knowledge is preferentially encoded in
or retrievable through $k_{local}$. To isolate proficiency, we measure the same
contrast on culture-agnostic content, the \emph{GlobalGap},

\begin{equation}
\text{GlobalGap}_j = \theta_{j,\text{CA},k_{local}} - \theta_{j,\text{CA},\text{en}},
\end{equation}
on which no local-knowledge advantage is expected.

Our second goal is to separate the two effects. Subtracting the GlobalGap
from the LocalGap cancels proficiency and isolates knowledge access; we call
this difference-in-differences contrast the \emph{KnowledgeGap},
\begin{equation}
\text{KnowledgeGap}_j=\text{LocalGap}_j-\text{GlobalGap}_j
\end{equation}
A positive KnowledgeGap indicates a local-language advantage on
culture-specific content beyond what proficiency explains.

\subsection{1PL-IRT Model}
\label{sec:metrics:crm}

In order to estimate model ability and task difficulty separately, we use 
a One-Parameter Logistic Item Response
Theory (1PL-IRT) model, equivalent to the Rasch model
\citep{rasch}. 
making abilities comparable across benchmarks of varying difficulty.
It models the probability that examinee $j$ answers item $i$ correctly
through the examinee's ability $\theta_j$ and the item difficulty
$b_i$. In our setting each logical item is prompted in multiple
languages, and we want difficulty to reflect the problem rather than
the prompt language. 

\begin{assumption}[Translation Equivalence]
\label{ass:translation_invariance}
    An underlying MCQ item has the same intrinsic difficulty across
    the languages it is translated into.
\end{assumption}

\begin{definition}[Multi-Facet Problem]
\label{def:multi_facet_problem}
    A single logical MCQ item $i \in I$ is prompted through a facet
    $k \in K$, i.e.\ a translated language. By
    Assumption~\ref{ass:translation_invariance}, the difficulty $b_i$
    is one scalar shared across all facets $k$.
\end{definition}

The residual variation in correctness across languages is then
absorbed by an ability specific to each subset
$s \in \{\text{CS}, \text{CA}\}$ and language $k$. Writing
$Y_{ijsk} \in \{0,1\}$ for the response of examinee $j$ to item $i$ in
subset $s$ and language $k$:

\begin{definition}[Facet-Conditioned 1PL-IRT Model]
\label{def:1plirt}
    \begin{equation}
        P(Y_{ijsk} = 1 \mid \theta_{jsk}, b_i)
        \coloneqq \sigma(\theta_{jsk} - b_i),
    \end{equation}
    with $\sigma$ the logistic function and $\theta_{jsk}$ the ability
    of $j$ on subset $s$ in language $k$.
\end{definition}

Assuming local independence, the model factorizes over items. Since
the facet enters through $\theta$ rather than $b$, the within-item
log-odds difference across two languages cancels difficulty,
\begin{equation}
    L_{ijsk} - L_{ijsk'} = \theta_{jsk} - \theta_{jsk'},
\end{equation}
making language contrasts difficulty-free --- the property our gap
metrics rely on. The full specification
appears in Appendix~\ref{app:irt}.

\begin{figure*}[h]
\centering
  \includegraphics[width=\linewidth]{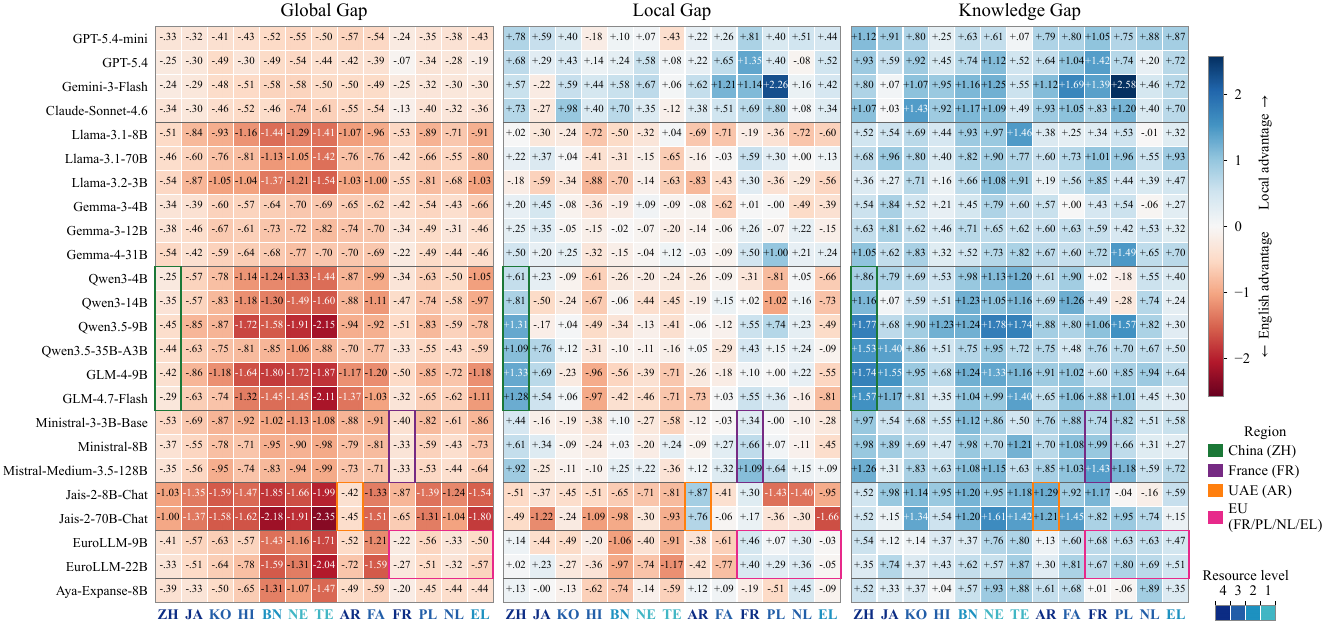}
  \caption{GlobalGap, LocalGap, and KnowledgeGap across models and locales, showing a consistent English proficiency advantage but a near-universal local-language knowledge advantage.}
  \label{fig:interactions_heatmap_main_figure}
\end{figure*}

\subsection{Dataset Construction and Languages}
\label{sec:method:data}
We cover 13 languages (locales) spanning diverse resource levels, scripts, and regions:
Chinese, Arabic, Greek, Hindi, Korean, French, Japanese, Bengali, Dutch,
Nepali, Persian, Polish, and Telugu. For each locale we build three matched
subsets: culture-agnostic items (CA), culture-specific items in the local
language (CS), and their English counterparts (CS\textsubscript{EN}). Each language is associated with a resource level described in \citep{joshi-etal-2020-state}. Appendix \ref{app:dataset} reports per-locale statistics.

\paragraph{Culture-agnostic (CA).}
We draw CA items from Global-MMLU, which provides high quality parallel translated questions across languages and human-assigned culture-agnostic labels; for each locale, we select the CA items in the corresponding language.

\paragraph{Culture-specific (CS).}
We source native local-language CS items from regional
benchmarks: CMMLU,
ArabicMMLU, GreekMMLU,
MILU, and KMMLU, with the remaining locales drawn from INCLUDE. From these we manually select subsets that are potentially cultural-related, and then ask the native speakers to select only genuinely culture-specific items, as described in Appendix~\ref{app:human}. 

\paragraph{English counterparts (CS\textsubscript{EN}).}
We translate each CS item into English with Claude Sonnet 4. We hired native speakers
to assess every item, rating translation quality and labeling whether the
question is culturally specific; we retain only high-quality translations
with genuine cultural content. CS and CS\textsubscript{EN} thus contain the
same logical questions in two query languages, isolating the effect of query
language on cultural knowledge access. Annotation details are in
Appendix~\ref{app:human}. 

\section{Experiments}
\label{sec:experiments}


\subsection{Models}
\label{sec:exp:models}
We evaluate roughly 80 models in three groups, testing both base and
instruction-tuned variants where available. Appendix \ref{app:dataset} lists all
models with its parameter count and country of origin.

\paragraph{Closed-source frontier.}
The proprietary models GPT-5.4, GPT-5.4-mini \citep{gpt5}, Gemini-3-Flash
\citep{gemini3}, and Claude-Sonnet-4.6 \citep{claudesonnet}.

\paragraph{Multilingual pretrained.}
Open-weight models pretrained from scratch on large, multilingual corpora,
spanning a range of countries of origin: Llama \citep{llama} and Gemma
\citep{gemma2, gemma3, gemma4} (USA), Qwen \citep{qwen25, qwen3, qwen35} and
GLM \citep{glm4, glm45} (China), Mistral \citep{magistral, ministral} (France),
Jais-2 \citep{jais} (UAE), EuroLLM \citep{eurollm} (EU), and Aya-Expanse
\citep{ayaexpense} (Canada).

\paragraph{Regionally-adapted.}
Models obtained by continued pretraining or fine-tuning of a global base on
local-language and cultural data, rather than pretrained from scratch:
AceGPT-v2 (Arabic) \citep{acegpt}, Llama-Krikri \citep{krikri} and Meltemi
\citep{meltemi} (Greek), PersianMind (Persian) \citep{persianmind}, and Bielik
(Polish) \citep{bielik}. 

\paragraph{Model filtering.}
Our decomposition assumes each model has enough multilingual competence for
the gaps to be meaningful. We therefore retain only models whose
accuracy exceeds $0.35$ on every subset, keeping the estimation within a
regime where the model demonstrably engages with each condition.

\subsection{Prompts and Inference}
\label{sec:exp:prompts}
\paragraph{Prompt templates.}
Each cell uses a fixed template rendered entirely in the query language, so no
cross-lingual cues leak in. CA questions use a \texttt{Question:
\{question+options\} Answer:} format; CS questions prepend a locale cue,
\texttt{In \{country\}, Question: \{question+options\} Answer:}, to ground them
in local context. Each template is then translated into the query language;
CS\_EN uses the same CS template in English without translation. We apply no
chat template and no reasoning/chain-of-thought prompting, consistent with our
log-likelihood scoring. For closed source models, we use free-generation and extract the answer option.

\paragraph{Inference.}
All open-weight models run in bfloat16 without quantization at temperature~0
with a fixed seed. Smaller models run on NVIDIA A6000 GPUs and larger models on
A100 80GB.

\subsection{IRT Fitting}
\label{sec:exp:crm}

At our benchmarking scale, classical estimators such as
Newton--Raphson or Expectation--Maximization are less practical. We
instead frame the facet-conditioned 1PL model as a continuous
optimization problem: treating the ability matrix $\boldsymbol{\Theta}$
and difficulty vector $\boldsymbol{B}$ as trainable embeddings, we
recover the maximum-likelihood estimates by minimizing the negative
log-likelihood, which coincides with the binary cross-entropy loss,
using the Adam optimizer \citep{adam} with mini-batch SGD, more details are in Appendix \ref{app:hyper_set}.

The 1PL likelihood is non-identifiable: since $\sigma$ depends only on
$\theta_{jsk} - b_i$, adding a constant to all abilities and all
difficulties leaves every probability unchanged. To guarantee a unique
solution we anchor the latent scale by constraining the item
difficulties to mean zero, applying the projection
$b_i \leftarrow b_i - \bar{b}$ and
$\theta_{jsk} \leftarrow \theta_{jsk} - \bar{b}$ (with
$\bar{b} = \frac{1}{|I|}\sum_{i \in I} b_i$) after each update. This
fixes the origin of the scale and gives a unique, interpretable
solution. The full derivation appears in
Appendix~\ref{app:irt-fitting}.

\section{Results}
\label{sec:results}
Figure \ref{fig:interactions_heatmap_main_figure} reports a representative subset of model results. For readability, we select models that are generally pretrained from scratch and cover diverse model families, regions of origin, and parameter scales. Regionally adapted models are discussed separately in the Section \ref{sec:ablations}. Full results for all tested models are provided in the Appendix \ref{app:full_results}.


\subsection{A Universal English Proficiency Advantage}
\label{sec:results-global}
Across the full grid, \textsc{GlobalGap} is negative in almost every cell
(Figure~\ref{fig:interactions_heatmap_main_figure}, left panel): for all models,
on all 13 languages, the estimated ability on culture-agnostic questions is
higher under English queries than under local-language queries. The cell-level
mean is $-0.79$, and no cell is positive. This pattern holds even for the
highest-resource non-English languages in our set, including Chinese ($-0.44$)
and French ($-0.37$), which still favor English on the CA subset. We read this
as primarily a proficiency baseline: largely independent of cultural content,
English yields higher estimated ability for these models, and the effect is
consistent enough that we treat it as a background against which the other two
gaps should be interpreted.

\subsection{The Local Language Tends to Win With High Resources or Matching Origin}
\label{sec:results-local}
Unlike \textsc{GlobalGap}, \textsc{LocalGap} is mixed
(Figure~\ref{fig:interactions_heatmap_main_figure}, center panel). The positive,
local-favoring cells concentrate on the highest-resource languages: Chinese has
the most positive column mean ($+0.48$), followed by French ($+0.42$), the two
clearest cases where querying in the local language yields higher estimated
ability on culture-specific questions than translating to English. For most
other languages \textsc{LocalGap} is negative or near zero, so on the CS ability
contrast translating to English is generally the better choice.

A second factor is model origin. Among models built in a given region, the
home-language \textsc{LocalGap} tends to be markedly higher than for other
models on the same language: $+0.79$ (Chinese-origin models on Chinese), $+0.94$
(UAE-origin on Arabic), and $+0.32$ (French-origin on French)
(Table~\ref{tab:home-language}). The Arabic case is notable because Arabic's
overall column mean is slightly negative, so here the advantage appears to come
from origin rather than resource level. The local language thus tends to win
when it is resource-rich, matched to the model's region of origin, or both.

\subsection{The Knowledge Advantage Is Universal but Driven by Origin, Not Resources}
\label{sec:results-knowledge}
\textsc{KnowledgeGap}, the knowledge-access component
(Figure~\ref{fig:interactions_heatmap_main_figure}, right panel), is positive in
98\% of cells: the knowledge component favors the local language almost
everywhere, including cells where the corresponding \textsc{LocalGap} is
negative.

This dissociation is the point of the decomposition: once proficiency is
partialled out, the local language carries a positive knowledge advantage.

Unlike the two performance gaps, \textsc{KnowledgeGap} is not explained by
resource level. What does track it is model origin: in-region models carry a
larger home-language \textsc{KnowledgeGap} than other models ($+0.70$ ZH,
$+0.60$ AR, $+0.30$ FR; Table~\ref{tab:home-language}), suggesting the advantage
is broadly present across models but is deepened by targeted regional training
rather than corpus scale.

\subsection{Performance Gaps Scale with Resource Level}
\label{sec:results-resource}
The English advantage is not uniform across languages but varies with resource
level. Table~\ref{tab:resource-groups} groups 13 languages into four ordinal
resource levels and reports the mean of each gap. Both \textsc{GlobalGap} and
\textsc{LocalGap} become less negative as resource level rises, moving from
$-1.22$ to $-0.54$ and from $-0.27$ to $+0.22$ respectively across the 13
per-language means, resource level correlates with both
(\textsc{GlobalGap}: Pearson $r=+0.84$, $p<0.001$; \textsc{LocalGap}: $r=+0.69$,
$p=0.009$). The \textsc{KnowledgeGap} shows no comparable trend ($r=-0.30$), a dissociation we return to in Section~\ref{sec:results-knowledge}. The
proficiency penalty thus shrinks as a language becomes higher-resource,
consistent with \textsc{GlobalGap} being driven primarily by how well the model
handles the language itself.

\begin{table}[h]
\centering
\small
\setlength{\tabcolsep}{4pt}
\begin{tabular}{c p{2.6cm} r r r}
\toprule
Level & Languages & Global & Local & Know. \\
\midrule
1 & TE, NE & $-1.22$ & $-0.27$ & $0.95$ \\
2 & BN, EL & $-0.95$ & $-0.26$ & $0.70$ \\
3 & HI, FA, KO, PL, NL & $-0.76$ & $-0.07$ & $0.68$ \\
4 & AR, JA, ZH, FR & $-0.54$ & $+0.22$ & $0.77$ \\
\bottomrule
\end{tabular}
\caption{Mean gaps by resource level, averaged over
models within each level group. Both \textsc{GlobalGap} and \textsc{LocalGap}
rise (become less negative) with resource level; \textsc{KnowledgeGap} does not
show a clear monotonic trend.}
\label{tab:resource-groups}
\end{table}

\subsection{Closed-Source Models Surface the Local-Language Advantage}
\label{sec:results-closed}
The four proprietary frontier models---GPT-5.4-mini, GPT-5.4, Gemini-3-Flash,
and Claude-Sonnet-4.6 answer most culture-specific
questions better in the local language than in English, with positive
\textsc{LocalGap} in most cells alongside positive \textsc{KnowledgeGap} in
nearly all of them. Because these models carry the smallest proficiency penalty
(\textsc{GlobalGap} among the least negative in the grid), the local-language
advantage is no longer masked and surfaces directly in the uncorrected
contrast---the same mechanism behind the Chinese result, here driven by model
capability rather than resource level. This also indicates the culture-specific
items genuinely require local knowledge.

\begin{table*}[t]
\centering
\small
\begin{tabular}{l c c c c c c}
\toprule
& \multicolumn{3}{c}{\textsc{LocalGap}} & \multicolumn{3}{c}{\textsc{KnowledgeGap}} \\
\cmidrule(lr){2-4}\cmidrule(lr){5-7}
Region $\to$ Lang & In & Out & $\Delta$ & In & Out & $\Delta$ \\
\midrule
China $\to$ ZH ($n{=}6$)  & $+1.07$ & $+0.28$ & $+0.79$ & $+1.44$ & $+0.74$ & $+0.70$ \\
UAE $\to$ AR ($n{=}2$)    & $+0.82$ & $-0.12$ & $+0.94$ & $+1.25$ & $+0.65$ & $+0.60$ \\
France $\to$ FR ($n{=}3$) & $+0.70$ & $+0.38$ & $+0.32$ & $+1.05$ & $+0.75$ & $+0.30$ \\
Europe $\to$ FR/PL/NL/EL ($n{=}2$) & $+0.23$ & $+0.05$ & $+0.18$ & $+0.64$ & $+0.64$ & $-0.01$ \\
\bottomrule
\end{tabular}
\caption{Home-language gaps for in-region models versus all other models, for
\textsc{LocalGap} and \textsc{KnowledgeGap}. In-region models show a larger
\textsc{LocalGap} on their home language in every case. The
\textsc{KnowledgeGap} boost holds for the single-language regions (China, UAE,
France).}
\label{tab:home-language}
\end{table*}

\section{Ablations}
\label{sec:ablations}

\subsection{Both Gaps Grow with Model Scale}
\label{sec:model-size}
\begin{figure}[h]
  \centering
  \includegraphics[width=\columnwidth]{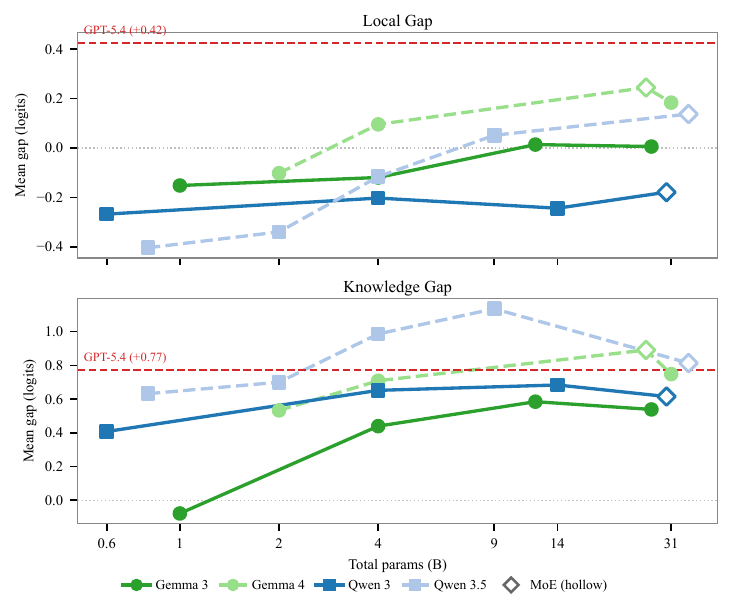}
  \caption{\textsc{LocalGap} (top) and \textsc{KnowledgeGap} (bottom) versus
  model size for four model families. The red dashed line marks the frontier reference GPT-5.4.}
  \label{fig:model_size_lines}
\end{figure}
To probe how the decomposition behaves with scale, we plot \textsc{LocalGap} and
\textsc{KnowledgeGap} against total parameter count for four model families
(Figure~\ref{fig:model_size_lines}). Both gaps tend to grow with size. The trend
is clearest for \textsc{LocalGap}, which rises within every family---most steeply
for the newer Gemma~4 and Qwen~3.5, gently for Gemma~3, and roughly flat for
Qwen~3---consistent with scale gradually reducing the proficiency penalty that
masks the local-language advantage.

\textsc{KnowledgeGap} also increases overall but not monotonically: some families
peak at intermediate sizes and then flatten or decline, most visibly Qwen~3.5
near 9B. The MoE checkpoints move inconsistently relative to dense models of
similar size, so we draw no general conclusion about MoE. The two metrics thus
scale differently---\textsc{LocalGap} rising steadily, \textsc{KnowledgeGap}
rising then leveling---but both persist across the full size range, mostly below
the GPT-5.4 reference, rather than being tied to any particular scale.

\begin{table*}[h]
\centering
\small
\setlength{\tabcolsep}{4.5pt}
\begin{tabular}{ll rr rr rr}
\toprule
 &  & \multicolumn{2}{c}{\textsc{Global}} & \multicolumn{2}{c}{\textsc{Local}} & \multicolumn{2}{c}{\textsc{Know.}} \\
\cmidrule(lr){3-4}\cmidrule(lr){5-6}\cmidrule(lr){7-8}
Model & Lg & Adapt. & $\Delta$ & Adapt. & $\Delta$ & Adapt. & $\Delta$ \\
\midrule
Krikri-8B          & EL & $-0.44$ & $+0.33$ & $+0.39$ & $+0.65$ & $+0.83$ & $+0.32$ \\
Meltemi-7B         & EL & $-0.47$ & $+0.30$ & $+0.48$ & $+0.74$ & $+0.94$ & $+0.44$ \\
Bielik-Minitron-7B    & PL & $-0.41$ & $+0.25$ & $+0.75$ & $+0.64$ & $+1.16$ & $+0.40$ \\
PersianMind        & FA & $-0.55$ & $+0.34$ & $+0.14$ & $+0.18$ & $+0.69$ & $-0.16$ \\
AceGPT-v2-8B       & AR & $-0.23$ & $+0.52$ & $+0.31$ & $+0.36$ & $+0.54$ & $-0.16$ \\
\midrule
Mean (Adapt.)      &    & $-0.42$ & $+0.35$ & $+0.41$ & $+0.52$ & $+0.83$ & $+0.17$ \\
\bottomrule
\end{tabular}
\caption{Language-adapted models evaluated on their target languages
\textit{Adapt.} denotes the adapted model's gap value. 
$\Delta$ denotes the difference between the adapted model and the per-language baseline, defined as the mean gap of general-purpose models evaluated on the same language.}
\label{tab:lang-adapt}
\end{table*}

\subsection{Language Adaptation Helps LocalGap but Not Necessarily KnowledgeGap}
\label{sec:lang-adapt}
We ask whether the masked advantage persists when a model is explicitly adapted
to a single target language. If the \textsc{LocalGap} penalty is mainly a
proficiency deficit, language-specific supervised fine-tuning (SFT) should reduce
or reverse it, whereas the \textsc{KnowledgeGap}---which reflects where cultural
knowledge is accessible rather than surface proficiency---need not move. We
evaluate five publicly available models adapted toward a non-English language,
each reported on its adapted language and compared against a \emph{per-language}
baseline: the average of the general pretrained models on that same
language. 

Table~\ref{tab:lang-adapt} shows two consistent effects and one that divides by
language. \textsc{LocalGap} turns positive for all five models, above a baseline
that is itself negative or near zero, and \textsc{GlobalGap} rises toward zero
for every model but stays negative; adaptation thus narrows the proficiency gap
to English without closing it, consistent with proficiency being what SFT most
readily moves. \textsc{KnowledgeGap}, by contrast, does not move uniformly: the
European models gain over their baseline while the Persian and Arabic models fall
slightly below it. This split suggests that SFT adds target-language cultural
knowledge for Greek and Polish, but for Persian and Arabic mainly improves
linguistic alignment, with little knowledge advantage beyond the same-language
general models.

\subsection{Latent Ability $\theta$ Recovers Raw Accuracy}
\label{sec:theta-acc}
Because $\theta$ is estimated from observed responses, we verify it faithfully
reflects raw accuracy, so that ability contrasts can be read as
difficulty-adjusted accuracy differences. Across all subsets, the mean Pearson
correlation between accuracy and $\theta$ is $0.996$, ranging narrowly from
$0.982$ (Greek CS) to $0.999$ (Arabic CS), and the Spearman correlation is
$0.9999$, indicating that $\theta$ recovers the examinee rank order almost
exactly. This is expected: within a subset, $\theta$ is by construction a
monotonic re-expression of accuracy on a continuous log-odds scale. As a result,
the within-component \textsc{GlobalGap} and \textsc{LocalGap} closely track the
corresponding local-minus-English accuracy differences; the IRT scale adds value by putting them on the same difficulty-adjusted footing as the cross-component \textsc{KnowledgeGap}, a comparison raw accuracy cannot support.

\subsection{Selection of the IRT Model}
\label{sec:irt-ablation}
We ablate two orthogonal choices in a $2\times2$ grid: the item-response
representation (binary vs.\ a Continuous Response Model, CRM, which scores
each item by the normalized probability mass over option tokens rather than a
hard label) and the inference algorithm (maximum likelihood, MLE, vs.\ the
No-U-Turn Sampler, NUTS); both are defined in Appendix~\ref{app:irt-variants}.
All four combinations are evaluated under 5-fold out-of-distribution
cross-validation on held-out responses, scored by predictive log-likelihood
(BCE), calibration (Brier), and discrimination (AUC, accuracy) (Table~\ref{tab:cv}).

\begin{table}[t]
\centering
\small
\setlength{\tabcolsep}{3.5pt}
\begin{tabular}{lccccc}
\toprule
Variant & BCE$\downarrow$ & SD & Brier$\downarrow$ & AUC$\uparrow$ & Acc$\uparrow$ \\
\midrule
binary+mle  & \textbf{0.513} & 9e-4 & \textbf{0.169} & \textbf{0.830} & \textbf{0.752} \\
binary+nuts & 0.515 & 1e-3 & 0.170 & 0.829 & 0.748 \\
crm+mle     & 0.537 & 5e-4 & 0.178 & 0.816 & 0.746 \\
crm+nuts    & 0.553 & 8e-4 & 0.182 & 0.821 & 0.746 \\
\bottomrule
\end{tabular}
\caption{Cross-validation performance across IRT variants. BCE and Brier measure predictive likelihood and calibration; AUC and
accuracy measure discrimination on held-out responses. A standard deviation (SD) is reported along with BCE for different folds.}
\label{tab:cv}
\end{table}

The binary representation outperforms CRM under both inference methods, and
MLE matches or slightly exceeds NUTS at far lower cost---likely because exact
posterior sampling is difficult in this high-dimensional parameter space. The
CRM gap is consistent with soft option-token probabilities being noisy and
poorly calibrated, so thresholding to binary correctness yields a more reliable
ability estimate. We therefore adopt \texttt{binary+mle}.

\section{Conclusion}
\label{sec:conclusion}
We introduced a controlled framework that combines a difference-in-differences design with item-response modeling to study language-conditioned access to localized knowledge in LLMs. By crossing question type, culture-agnostic versus culture-specific, with query language, English versus local, we separate general language proficiency from localized knowledge access. Specifically, we estimate \textsc{GlobalGap} as the proficiency gap, \textsc{LocalGap} as the combined effect on culture-specific questions, and \textsc{KnowledgeGap} as their difference on a shared, difficulty-adjusted 1PL IRT scale.

Across 13 locales and roughly 80 models, \textsc{LocalGap} is shaped by language resource level, model origin, and language adaptation, becoming positive mainly when these factors favor the local language. In contrast, \textsc{KnowledgeGap} is positive in nearly all settings, suggesting that local languages generally provide better access to local cultural knowledge, even when this advantage is hidden by weaker local-language proficiency.

These findings offer useful insight for current
localized-knowledge benchmarks and language-adaptation
strategies. When a model performs worse in a local
language, the cause is often not missing cultural
knowledge but weaker proficiency in that language, which
masks a knowledge advantage.
Improving proficiency---through language adaptation or
stronger multilingual training---is therefore a possible
way to unlock this hidden capability, since the knowledge
access largely exists and is held back mainly by the
proficiency penalty.

\newpage
\section*{Limitations}

We note several limitations of this study. First, the
size of the culture-specific subsets varies across
locales, in part because some languages do not have
large-scale cultural benchmark data readily available
online. 

Second, our evaluation of regionally adapted models is
limited in scope. Strong open-weight models that have been explicitly adapted to a non-English language remain relatively scarce, and high-quality, widely-adopted models of this kind are rarer still. So we were able to include only a small
number of such models, covering a handful of languages.

Third, all of our results are based on multiple-choice
questions scored by log-likelihood. We do not evaluate
open-ended generation, where knowledge access may behave
differently and where failures can take more diverse forms
than a fixed set of options allows. 


 
\bibliography{small_acl,custom}

\appendix
\newpage
\section{Model and Dataset Details}
\label{app:dataset}

We gathered 79 different models with various types, sizes and origins as the population of the benchmarking effort, as further details are shown in Table~\ref{tab:models}. Both closed-source and open-source models are evaluated, with multiple regionally adapted LLMs considered. In the meantime, both culture-specific(CS) and culture-agnostic (CA) benchmarks are constructed for 14 languages from different language families with spanning richness in resources, as illustrated in Table~\ref{tab:language-overview}. Multiple languages with extremely low resources are included in particular to study the full spectrum of the knowledge gap of different pretraining scenarios.  

\begin{table*}[htbp]
\centering
\small
\setlength{\tabcolsep}{5pt}
\begin{tabular}{@{}lll r@{}}
\toprule
\textbf{Model Family} & \textbf{Origin} & \textbf{Parameter Sizes} & \textbf{\#} \\
\midrule
\multicolumn{4}{@{}l}{\textsc{Closed Source}} \\
\addlinespace[2pt]
GPT-5.4            & OpenAI, USA        & ---                                & 2 \\
Gemini 3 Flash     & Google, USA        & ---                                & 1 \\
Claude Sonnet 4.6  & Anthropic, USA     & ---                                & 1 \\
\midrule
\multicolumn{4}{@{}l}{\textsc{Multilingual Pretrained}} \\
\addlinespace[2pt]
Llama 3.1          & Meta, USA          & 8B, 70B                            & 4 \\
Llama 3.2          & Meta, USA          & 1B, 3B                             & 4 \\
Gemma 2            & Google, USA        & 9B, 27B                            & 4 \\
Gemma 3            & Google, USA        & 1B, 4B, 12B, 27B                  & 8 \\
Gemma 4            & Google, USA        & E2B, E4B, 26B-A4B, 31B            & 8 \\
Qwen 2.5           & Alibaba, China     & 7B, 14B, 32B                       & 6 \\
Qwen 3             & Alibaba, China     & 0.6B, 1.7B, 4B, 14B, 30B-A3B     & 7 \\
Qwen 3.5           & Alibaba, China     & 0.8B, 2B, 4B, 9B, 27B, 35B-A3B   & 11 \\
GLM 4 / 4.7        & Zhipu AI, China    & 9B                                 & 3 \\
Mistral            & Mistral AI, France  & 119B, 128B                         & 2 \\
Ministral 3        & Mistral AI, France  & 3B, 8B, 14B                        & 6 \\
Jais 2             & Inception, UAE      & 8B, 70B                            & 2 \\
EuroLLM            & Europe              & 9B, 22B                            & 4 \\
Aya Expanse        & Cohere, Canada      & 8B                                 & 1 \\
\midrule
\multicolumn{4}{@{}l}{\textsc{Regionally Adapted}} \\
\addlinespace[2pt]
AceGPT-v2          & China (Arabic focus)               & 8B                                 & 1 \\
Llama-Krikri       & Greece              & 8B                                 & 1 \\
Meltemi            & Greece              & 7B                                 & 1 \\
Bielik-Minitron    & Poland              & 7B                                 & 1 \\
PersianMind        & Iran                & ---                                & 1 \\
\midrule
\multicolumn{3}{@{}l}{\textbf{Total}} & \textbf{79} \\
\bottomrule
\end{tabular}
\caption{Overview of models evaluated in our benchmark, grouped by family. Both base and instruction-tuned variants are included where available;
\textbf{\#} denotes the total number of model variants per family.}
\label{tab:models}
\end{table*}

\begin{table*}[htbp]
\centering
\small
\begin{tabular}{lllcrrl}
\toprule
Language & Code & Family & Resource & \#CA & \#CS & Source \\
\midrule
English & en & Indo-European & 5 & 2,058 & 6,093 & --- \\
Chinese & zh & Sino-Tibetan & 4 & 2,058 & 1,000 & CMMLU \\
Arabic & ar & Afro-Asiatic & 4 & 2,058 & 999 & ArabicMMLU \\
French & fr & Indo-European & 4 & 2,058 & 139 & INCLUDE \\
Japanese & ja & Japonic & 4 & 2,058 & 73 & INCLUDE \\
\addlinespace
Dutch & nl & Indo-European & 3 & 2,058 & 136 & INCLUDE \\
Hindi & hi & Indo-European & 3 & 2,058 & 1,000 & MILU \\
Korean & ko & Koreanic & 3 & 2,058 & 1,000 & KMMLU \\
Persian & fa & Indo-European & 3 & 2,058 & 185 & INCLUDE \\
Polish & pl & Indo-European & 3 & 2,058 & 75 & INCLUDE \\
\addlinespace
Bengali & bn & Indo-European & 2 & 2,058 & 342 & INCLUDE \\
Greek & el & Indo-European & 2 & 2,058 & 785 & GreekMMLU \\
\addlinespace
Nepali & ne & Indo-European & 1 & 2,058 & 230 & INCLUDE \\
Telugu & te & Dravidian & 1 & 2,058 & 129 & INCLUDE \\
\midrule
\textbf{Total} & \multicolumn{2}{l}{14 languages} & & \textbf{28,812} & \textbf{12,186} & \\
\bottomrule
\end{tabular}
\caption{Overview of languages and subsets used in our study. The CA (culture-agnostic) subset is sourced from GlobalMMLU. The CS (culture-specific) subset contains locally sourced questions probing culture-specific knowledge; CS items are also available in English translation (CS-en). All subsets are evaluated across 79 LLMs.}
\label{tab:language-overview}
\end{table*}

\section{Human evaluation}
\label{app:human}

\begin{tcolorbox}[
  colback=gray!5,
  colframe=black,
  boxrule=0.8pt,
  arc=2mm,
  width=\linewidth
]
\small

\textbf{Annotation Instructions}

This project evaluates the cultural ability of LLMs using questions in both the native language and English translation.

You will be given a CSV file containing multiple-choice questions with four options: A, B, C, and D. For each row, please complete two tasks.

Task 1: Translation Quality Score

Evaluate the English translation of the question and all answer options. Enter a score in the SCORE column.

1 – Poor: The translation loses important meaning, mistranslates key or culturally specific terms, or is misleading.

2 – Acceptable: The translation has minor ambiguity or awkwardness, but the overall intent is understandable.

3 – Good: The translation faithfully preserves the original meaning, and the question can be answered correctly from the English version alone.

Task 2: Local Culture Annotation

Indicate whether the question requires local cultural knowledge to answer correctly.

Yes: The question depends on knowledge of the target locale, such as local history, traditions, festivals, geography, laws, institutions, education, literature, religion, food, social norms, or popular culture.

No: The question can be answered using general knowledge and does not rely on local cultural context.

Unsure: The cultural relevance is unclear, weak, or ambiguous.

Please judge each row based on the question and answer options, not on whether you personally know the answer.

\end{tcolorbox}

\noindent\begin{minipage}{\linewidth}
\captionof{figure}{The annotation instruction for human native-language-speaking evaluators on translation quality control.}\label{fig:prompt}
\end{minipage}

\begin{table}[htbp]
\centering
\small
\begin{tabular}{lrlr}
\toprule
Language & Error (\%) & Language & Error (\%) \\
\midrule
Chinese   & 2.4 & Dutch    & 0.4 \\
Arabic    & 1.8 & Korean   & 0.9 \\
French    & 2.0 & Greek    & 0.1 \\
Japanese  & 4.0 & Telugu   & 0.0 \\
Hindi     & 1.6 & Bengali  & 2.3 \\
Persian   & 5.0 & Nepali   & 7.0 \\
Polish    & 1.7 &          &     \\
\bottomrule
\end{tabular}
\caption{Error rate (\%) in human evaluation of translation quality.}
\label{tab:translation_error}
\end{table}

To confirm the Assumption~\ref{ass:translation_invariance} for our IRT model, we asked native-speaking translators to verify the quality of the translations for the CS (culture-specific) items. Since every MCQ item originally written in a native language is automatically translated into English, we evaluate each item individually for correctness, rigor, and topic relevance. 

For each language that is not the native language among the authors, we hired native speakers to do the annotation, they are paid 15 dollars per hour, and each subset requires around 2 hours of work. Each annotator is given an annotation instruction as described in Figure \ref{fig:prompt}.

Table~\ref{tab:translation_error} shows that raw translation quality evaluated by human annotators, most local languages results in negligible error rates (translation scored 1), indicating that translation equivalence is well preserved. Items flagged with a score of 1 or 2 are filtered out, ensuring that the remaining benchmark is fully legitimate.

\section{Hyperparameters and Setup}
\label{app:hyper_set}

The 1PL-IRT model with MLE SGD is trained on a NVIDIA RTX A6000 48GB GPU. The training with learning rate 0.05 spans for about 600 to 1000 steps until convergence of the negative log-likelihood loss, taking no more than 1 minute. For ablations, the NUTS fitting process takes 8 sampling chains in parallel, each takes 2000 warmup steps followed by another 2000 sampling steps. For cross validation, all observed responses are partitioned completely randomly into five folds to provide a generalizable held-out cross validation setup. 

\section{Full Specification of the 1PL-IRT Model and Fitting}
\label{app:irt}

This appendix gives the complete specification of the
facet-conditioned 1PL-IRT model (\S\ref{sec:metrics:crm}) and the
fitting procedure (\S\ref{sec:exp:crm}), including the assumptions,
ability parameterization, identifiability argument, and anchoring
projection summarized in the main text.

\subsection{Model Details}
\label{app:irt-model}

The mathematical validity and interpretability of the facet-conditioned 1PL model rest upon three foundational assumptions concerning the joint distribution of responses and the latent space. 

\begin{assumption}[Unidimensionality]
    The latent trait space is one-dimensional, meaning that a single scalar value $\theta_{jsk}$ is strictly sufficient to explain the statistical dependence between item responses for subject $j$ on a subset $s$ with a certain prompt language $k$. 
\end{assumption}

\begin{assumption}[Local Independence]
\label{ass:local_independence}
    Conditional on the latent trait $\theta_{jsk}$, the responses to multi-facet problems are mutually independent. For a subset $s$ of size $n$ with prompt language $k$, the joint probability of a response vector $\boldsymbol{Y}_{jsk}=(y_{1jsk},y_{2jsk},\dots,y_{njsk})$ is the product of the individual marginal probabilities: 
    \begin{equation}
    \begin{aligned}
        P&(\boldsymbol{Y}_{jsk}=\boldsymbol{y}_{jsk}|\theta_{jsk},\boldsymbol{b}_s)\\
        &=\prod_{i=1}^nP(Y_{ijsk}=y_{ijsk}|\theta_{jsk},b_{is})
    \end{aligned}
    \end{equation}
\end{assumption}

\begin{assumption}[Monotonicity]
    The probability of a correct response $P(Y_{ijsk}=1|\theta_{jsk},b_i)$ is a strictly monotonically increasing function of the latent trait $\theta_{jsk}$, meaning that for any two examinees with latent traits $\theta_{1sk}, \theta_{2sk}\in\mathbb{R}$:
    \begin{equation}
        \theta_{1sk} < \theta_{2sk} \Leftrightarrow P_i(\theta_{1sk})<P_i(\theta_{2sk}),\forall i
    \end{equation}
\end{assumption}


\paragraph{Difficulty-free language contrasts.}
By treating the language facet as a dimension of the examinee's latent
space rather than a shift in item difficulty, the log-odds $L_{ijsk}$
of a correct response for examinee $j$ on item $i$ from subset $s$ in
language $k$ is
\begin{equation}
    L_{ijsk}
    = \ln\!\left(\frac{P(Y_{ijsk}=1)}{1 - P(Y_{ijsk}=1)}\right)
    = \theta_{jsk} - b_i,
\end{equation}
and likewise $L_{ijsk'} = \theta_{jsk'} - b_i$ for a second language
$k'$. The difficulty term cancels in the within-item difference,
\begin{equation}
    L_{ijsk} - L_{ijsk'} = \theta_{jsk} - \theta_{jsk'},
\end{equation}
so the language contrast for the same item is exactly the difference
in the examinee's language-specific abilities, independent of $b_i$.

\subsection{Fitting Procedure}
\label{app:irt-fitting}

At our benchmarking scale, classical estimators such as
Newton--Raphson or Expectation--Maximization are less practical. We
instead frame the facet-conditioned 1PL model as a continuous
optimization problem, treating the ability matrix $\boldsymbol{\Theta}$
of all facet-specific examinee abilities and the difficulty vector
$\boldsymbol{B}$ of all item difficulties as trainable embeddings and
recovering the maximum-likelihood estimates by stochastic gradient
descent (SGD).

Let $\boldsymbol{Y} = \{(j, i, s, k, y_{ijsk})\}$ denote the observed
benchmarking data, where each entry records examinee $j$, a
multi-facet MCQ item $i$ from subset $s$ prompted in language $k$, and
the dichotomous outcome $y_{ijsk} \in \{0, 1\}$. Writing $P_{ijsk}$ for
the logistic probability of Definition~\ref{def:1plirt} and invoking
Assumption~\ref{ass:local_independence}, the joint likelihood is the
product of the individual Bernoulli probabilities:
\begin{equation}
\begin{aligned}
    L&(\boldsymbol{\Theta}, \boldsymbol{B} \mid \boldsymbol{Y})\\
    &= \prod_{(j,i,s,k,y_{ijsk}) \in \boldsymbol{Y}}
      P_{ijsk}^{\,y_{ijsk}}
      (1 - P_{ijsk})^{\,1 - y_{ijsk}}.
\end{aligned}
\end{equation}

Taking the natural logarithm yields the log-likelihood $\ell$, and the
empirical loss $\mathcal{L}$ is the negative log-likelihood (NLL),
which coincides with the binary cross-entropy (BCE) loss of standard
machine-learning toolkits:
\begin{equation}
    \mathcal{L}(\boldsymbol{\Theta}, \boldsymbol{B})
    = -\ell(\boldsymbol{\Theta}, \boldsymbol{B})
    = -\log L(\boldsymbol{\Theta}, \boldsymbol{B} \mid \boldsymbol{Y}).
\end{equation}
We minimize $\mathcal{L}$ over $\boldsymbol{\Theta}$ and
$\boldsymbol{B}$ jointly using the Adam optimizer \citep{adam} with
mini-batch SGD.

\paragraph{Identifiability and anchoring.}
The 1PL model is non-identifiable: because the logistic function
depends only on the difference $\theta_{jsk} - b_i$, adding any
constant $c$ to all abilities and all difficulties leaves every
probability unchanged,
\begin{equation}
    \sigma\big((\theta_{jsk} + c) - (b_i + c)\big)
    = \sigma(\theta_{jsk} - b_i).
\end{equation}
To guarantee a unique MLE, we anchor the latent scale by constraining
the mean item difficulty to zero. After each SGD update we apply the
projection, with $\bar{b} = \frac{1}{|I|}\sum_{i \in I} b_i$,
\begin{equation}
    \begin{aligned}
        b_i &\leftarrow b_i - \bar{b}, && \forall i \in I, \\
        \theta_{jsk} &\leftarrow \theta_{jsk} - \bar{b},
        && \forall j, s, k.
    \end{aligned}
\end{equation}
This re-centering fixes the origin of the latent scale and gives a
unique, interpretable solution.

\subsection{IRT Variant Details}
\label{app:irt-variants}

\paragraph{Item-response representations.}
The \textbf{dichotomous} (binary) model scores each response as correct or incorrect ($y\in\{0,1\}$) and models $P_i(\theta_j)=\sigma(\theta_j-b_i)$ as in the main text (\S\ref{sec:irt-ablation}). The \textbf{Continuous Response Model (CRM)} instead treats the response as graded: rather than thresholding, it uses the probability mass the model places on the correct option after normalizing over the answer-option tokens, giving a continuous score in $[0,1]$. This retains finer information about how confident the model was, at the cost of inheriting any miscalibration in the raw token probabilities.

\paragraph{Inference algorithms.}
\textbf{MLE} treats $\theta$ and $b$ as free parameters with no prior and minimizes the negative log-likelihood by gradient descent (Adam), returning a single point estimate $(\hat\theta,\hat b)$. \textbf{NUTS} is an adaptive variant of Hamiltonian Monte Carlo that draws samples from the full Bayesian posterior over $(\theta,b)$; it is a standard choice for fitting IRT models and yields credible intervals directly, but its cost grows with the dimensionality of the parameter space and it can suffer from divergences and inefficient exploration in very high-dimensional settings.

\section{Full results}
\label{app:full_results}
\begin{figure*}[h]
\centering
  \includegraphics[width=\linewidth]{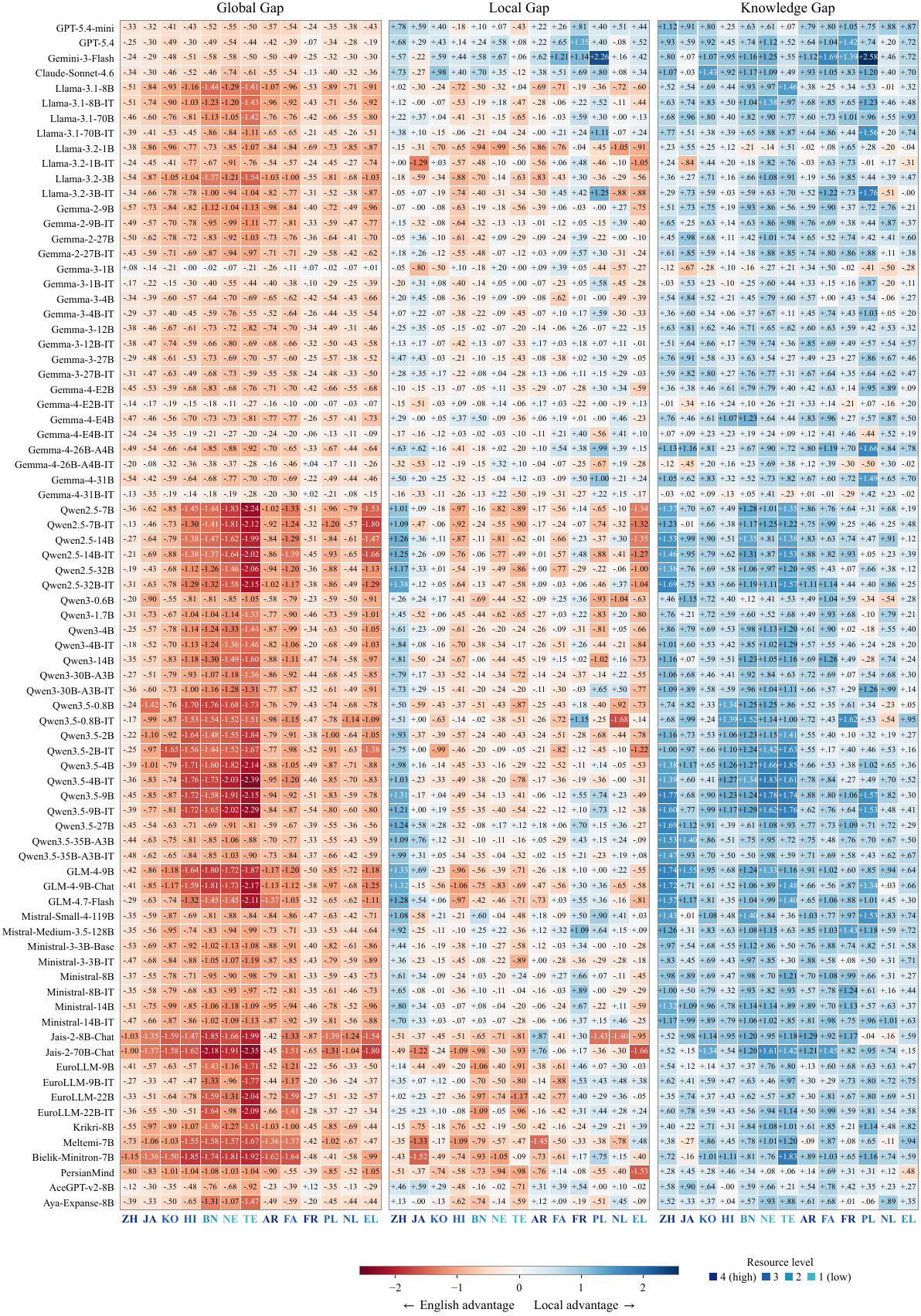}
  \caption{GlobalGap, LocalGap, and KnowledgeGap across all models and locales.}
  \label{fig:interactions_heatmap_default}
\end{figure*}

We present results for all models tested in Figure \ref{fig:interactions_heatmap_default}.

\end{document}